%% file: top.tex
\documentclass[conference, 9pt]{IEEEtran}
\IEEEoverridecommandlockouts
\usepackage{cite}
\usepackage{amsmath,amssymb,amsfonts}
\usepackage{algorithmic}
\usepackage{booktabs}
\usepackage{graphicx}
\usepackage{textcomp}
\usepackage{xcolor}
\usepackage{multirow}
\usepackage{algorithm}
\usepackage{multirow}
\usepackage{amsmath}
\usepackage{hyperref}
\usepackage{balance}
\usepackage{threeparttable}
\def\BibTeX{{\rm B\kern-.05em{\sc i\kern-.025em b}\kern-.08em
    T\kern-.1667em\lower.7ex\hbox{E}\kern-.125emX}}

\begin{document}

\title{DeepCell: Self-Supervised Multiview Fusion for \\Circuit Representation Learning}


\author{
	\IEEEauthorblockN{
		Zhengyuan Shi$^{1,5}$\textsuperscript{\textsection},
		Chengyu Ma$^{2,5}$\textsuperscript{\textsection},
		Ziyang Zheng$^{1,5}$, 
		Lingfeng Zhou$^{3,5}$, 
         Hongyang Pan$^{4,5}$,
        Wentao Jiang$^{2,5}$, \\
        Fan Yang$^4$, 
        Xiaoyan Yang$^3$, 
        Zhufei Chu$^2$ 
        and Qiang Xu$^{1,5}$} 
\IEEEauthorblockA{$^1$\textit{Department of Computer Science and Engineering}, \textit{The Chinese University of Hong Kong}, Sha Tin, Hong Kong S.A.R.\\}
\IEEEauthorblockA{$^2$\textit{Faculty of Electrical Engineering and Computer Science}, \textit{Ningbo University}, Ningbo, China \\}
\IEEEauthorblockA{$^3$\textit{School of Computer Science}, \textit{Hangzhou Dianzi University}, Hangzhou, China \\}
\IEEEauthorblockA{$^4$\textit{School of Microelectronics, State Key Laboratory of Integrated Chips and System}, \textit{Fudan University}, Shanghai, China\\}
\IEEEauthorblockA{$^5$\textit{National Center of Technology Innovation for EDA}, Nanjing, China \\}
}

\maketitle

\footnotetext{Both authors contributed equally to this research.}

\begin{abstract}
We introduce \textbf{DeepCell}, a novel circuit representation learning framework that effectively integrates multiview information from both And-Inverter Graphs (AIGs) and Post-Mapping (PM) netlists. At its core, DeepCell employs a self-supervised Mask Circuit Modeling (MCM) strategy, inspired by masked language modeling, to fuse complementary circuit representations from different design stages into unified and rich embeddings. To our knowledge, DeepCell is the first framework explicitly designed for PM netlist representation learning, setting new benchmarks in both predictive accuracy and reconstruction quality. We demonstrate the practical efficacy of DeepCell by applying it to critical EDA tasks such as functional Engineering Change Orders (ECO) and technology mapping. Extensive experimental results show that DeepCell significantly surpasses state-of-the-art open-source EDA tools in efficiency and performance. 

\end{abstract}

\input{01_introduction}
\input{02_related}

\input{03_method}
\input{04_experiment}

\input{041_ECO}
\input{042_Mapping}

\input{05_conclusion}

\balance


\bibliographystyle{IEEEtran}
\bibliography{reference}

\end{document}

%% file: 01_introduction.tex
\section{Introduction} \label{Sec:Intro}
Developing Integrated Circuits (ICs) is a highly intricate process that implements complex Boolean logic using billions of transistors, and then precisely places them on a tiny silicon wafer. To manage this complexity, Electronic Design Automation (EDA) tools break down the chip design process into distinct stages, each addressing different representations or modalities of the circuit. These modalities—such as Verilog code, And-Inverter Graphs (AIGs), Post-Mapping (PM) netlists, floorplans, and layouts—capture various aspects of the design at different points in the flow. 
As such, the EDA process is often viewed as a multimodal transformation~\cite{chen2024large, li2025deepcircuitx}, where multiple views of the circuit are generated and refined throughout the design cycle. 

While emerging circuit learning models~\cite{li2022deepgate, wang2022functionality, shi2023deepgate2, wu2023gamora, deng2024less, shi2024deepgate3, khan2024deepseq, liu2024polargate} have made significant strides in learning representations for individual circuit modalities, they often overlook the multiview information inherent in the design process. As a result, these techniques tend to focus on isolated design tasks at individual stages, such as design-for-test~\cite{shi2022deeptpi}, logic optimization~\cite{neto2019lsoracle} and functional reasoning~\cite{wu2023gamora, deng2024less}, without considering the more diverse spectrum of circuit modalities that could enhance their performance and generalization ability. Although recent cross-modal circuit learning approaches have successfully predicted Power, Performance and Area (PPA) metrics based on the early-stage designs~\cite{yao2024rtlrewriter, fang2025circuitfusion}, they still face significant challenges in automation and scalability when applied to cross-stage EDA tasks. These tasks, such as logic synthesis, technology mapping and engineering change, requiring integrated insights from multiple modalities remains unexplored. Consequently, \textit{how to efficiently leverage the multiview nature of the EDA flow into EDA tools is still an open question}. 

To answer such critical question, we introduce \textbf{DeepCell}, a self-supervised multiview fusion framework for learning netlist representation. DeepCell integrates information from both pre-mapping (i.e. AIGs) and post-mapping (i.e. PM netlists) modalities, using the corresponding Graph Neural Network (GNN)-based encoders. To fuse these multiview representation, we employ a self-supervised \text{Mask Circuit Modeling (MCM)} pretraining task, inspired by Masked Language Modeling (MLM)~\cite{devlin2018bert}, which leverages embeddings from one view to reconstruct embeddings of another. By bridging the gap between technology-independent and technology-dependent circuit representations using Transformer, DeepCell generates rich, generalizable and multiview embeddings for both AIGs and PM netlists. 

We validate DeepCell by applying it to two representative and practical EDA tasks: 1) Functional Engineering Change Orders (ECO): a PM netlist is patched to match a given golden design, avoiding the time-consuming process of re-implementation. 2) Technology mapping: an AIG is converted into PM netlist using a specific technology library. 
While these tasks may initially seem unrelated, both require a deep understanding of the two circuit views, a challenge that previous circuit learning solutions struggle to address efficiently. 
DeepCell, with its ability to capture and fuse multiview information, can be seamlessly integrated into existing EDA tools with minimal engineering effort and tiny runtime overhead. In the ECO task, DeepCell significantly reduces both engineering change costs and overall runtime compared to the champion solution~\cite{dao2018efficient} of the ICCAD Contest. Besides, DeepCell achieves lower area and delay than the state-of-the-art technology mapping engine in the open-source synthesis tool ABC~\cite{brayton2010abc}, demonstrating its superior performance and efficiency. 

Our key contributions are summarized as follows:
\begin{itemize}
    \item We propose DeepCell, a multiview representation learning and fusion framework for circuit netlists, incorporating the information from both technology-independent and technology-dependent circuit designs. 
    \item We introduce the first representation learning model tailored for PM netlist, employing a self-supervised pretraining mechanism to refine PM netlist embeddings using AIG representations.
    \item We demonstrate the effectiveness and generalization ability of DeepCell in various cross-modal EDA tasks, such as functional ECO and technology mapping. DeepCell can be seamlessly integrated as a plug-in into existing EDA tools, achieving significant efficiency improvements over state-of-the-art solutions. 
\end{itemize}

The remainder of this paper is organized as follows: Section~\ref{Sec:Related} reviews related work. Section~\ref{Sec:Method} describes the proposed DeepCell framework, including architecture and pretraining mechanism. Section~\ref{Sec:Experiment} presents the pretraining results and investigate the effect of proposed training strategy. Next, we apply DeepCell in functional ECO (see Section~\ref{Sec:ECO}) and technology mapping (see Section~\ref{Sec:Mapping}). Final Section~\ref{Sec:Conclusion} concludes this paper.

\begin{figure*}[!t]
    \centering
    \includegraphics[width=0.9\linewidth]{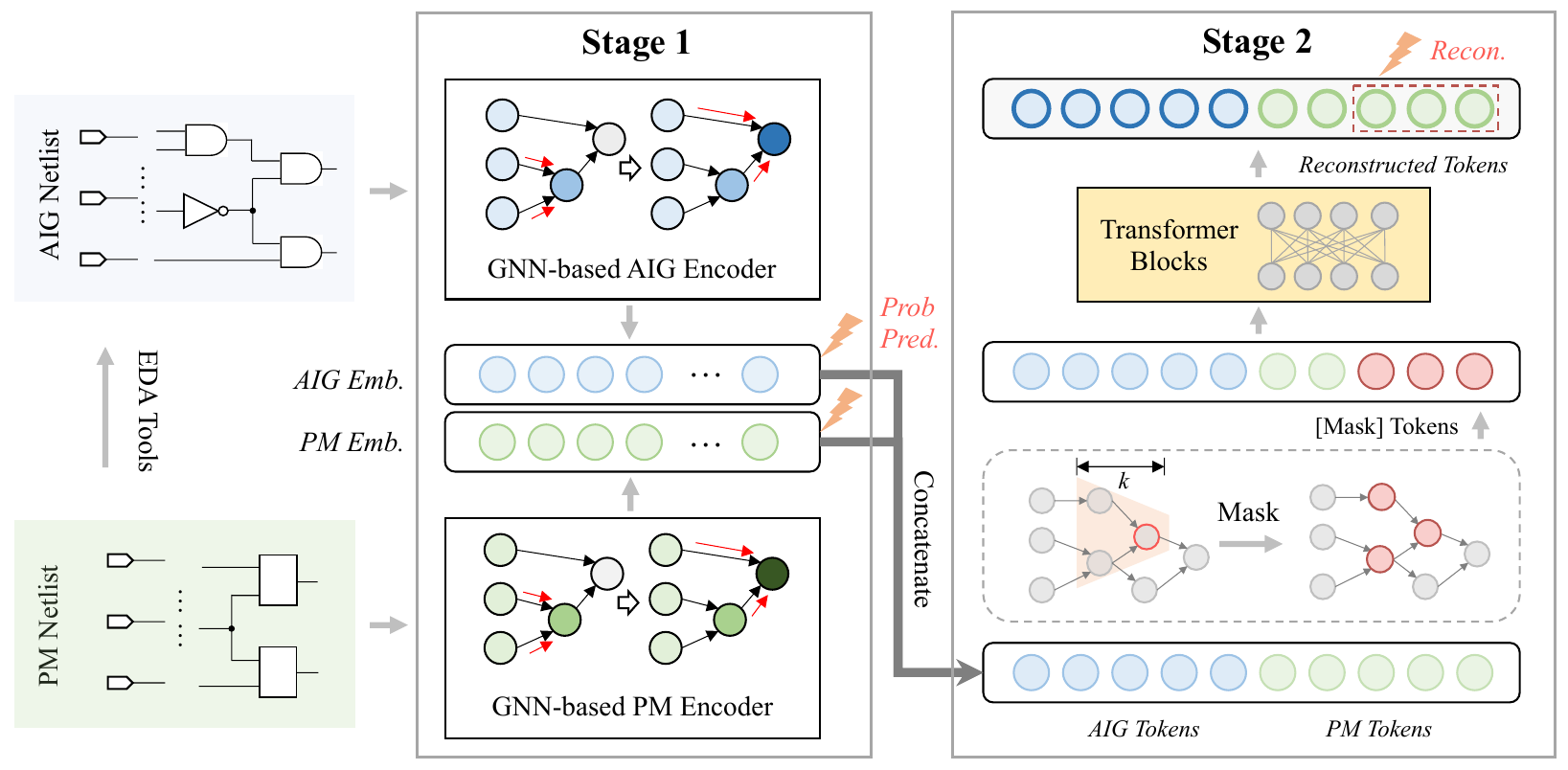}
    \caption{Overview of DeepCell illustrating the multiview fusion of AIG and PM netlist embeddings}
    \label{fig:overview}
    \vspace{-5pt}
\end{figure*}

%% file: 02_related.tex
\section{Related Work} \label{Sec:Related}
\subsection{Circuit Representation Learning}
Circuit representation learning has emerged as an attractive direction in the field of EDA, focusing on training models to obtain general circuit embeddings that can be applied to various downstream tasks~\cite{chen2024large}. The first circuit representation learning framework, DeepGate~\cite{li2022deepgate}, proposes supervising model training using logic-1 probability under random simulation and 
achieves substantial improvements in tasks like testability analysis~\cite{shi2022deeptpi} and logic equivalence checking~\cite{shi2024eda}.  Its successor, DeepGate2 \cite{shi2023deepgate2}, further refines this approach by separating functional and structural embeddings for different applications. Additionally, Gamora~\cite{wu2023gamora} and HOGA \cite{deng2024less} leverage sub-circuit identification as pre-training tasks, while FGNN \cite{wang2022functionality} trains models in an unsupervised contrastive manner by distinguishing between equivalent and non-equivalent circuits. DeepGate3~\cite{shi2024deepgate3} and DeepGate4~\cite{zheng2025deepgate4} introduce the framework that combines Graph Neural Networks (GNNs) and Transformer blocks to address the scalability challenges in circuit representation learning, allowing the model to improve with the growth of the training dataset. 

While these methods are significant, they primarily focus on single-modal circuit representations (e.g., AIGs), thus limiting their ability to exploit complementary information across multiple circuit modalities. Recent works~\cite{yao2024rtlrewriter, fang2025circuitfusion} improve large language models (LLMs) by enhancing the understanding of Verilog code using textual summaries and netlists, but they are not directly compatible with practical EDA tools for circuit-based applications. 
In contrast, DeepCell framework combines the information from both AIG and PM netlist, fusing circuit representations from different circuit instances. As a result, it stands out in its ability to handle critical EDA tasks that previous approaches struggle to address effectively due to their lack of multiview perception.




\vspace{-3pt}
\subsection{Functional ECO}
ECO are a critical component in the VLSI design process, used to rectify design problems after tape-out or chip fabrication. ECO involves making modifications to correct these errors, and they are indispensable in avoiding the high expenses associated with design re-spin ~\cite{jaeger2007virtually}.
For functional ECO, the purpose is to generate patch so that the given circuit is equal to the golden circuit, while minimizing the resource cost of the generated patches and making the running time as short as possible. Synthesis-based ECO algorithms are good at solving this problem~\cite{huang2013match}. It relies on a diagnostic strategy to identify internal rectifier signals, and then applies a resynthesis technique to generate patch functions for functional differences. These algorithms have been able to automate the process of functional ECO. However, their effectiveness depends heavily on the quality of the candidate signals for patch generation. When the candidate set is large, the process of selecting rectification cones often results in significant time consumption due to repeated invocations of satisfiability (SAT) and quantified Boolean formula (QBF) solving~\cite{jiang2020engineering}. Recent approaches attempt to mitigate this issue by prioritizing candidates based on patch cost metrics alone~\cite{kravets2020learning, liu2023feep}. Yet, these methods overlook driving relations between signals, lead to the incomplete search space exploration. 
Our DeepCell framework, with its multiview perception, can predict the driven logic of a given cell under the golden implementation. This ability allows DeepCell to reduce the candidate signal set, thereby enhancing the efficiency of ECO tools.



\vspace{-3pt}
\subsection{Technology Mapping}
Technology mapping is a critical step in logic synthesis, where an optimized Boolean network is transformed into a circuit composed of primitive elements supported by the target technology library. Currently, the most widely used approach for technology mapping is cut-based technology mapping~\cite{chatterjee2006reducing}. This method involves enumerating a set of structural cuts for each node in the network, computing the Boolean function of each cut, and matching these functions with the supergates available in the technology library. Matches are sorted based on a target heuristic, such as area, delay, or power consumption, and the optimal cover is selected during a graph traversal~\cite{costamagna2024enhanced}. However, this method relies on a load-independent abstraction of delay information, which approximates the delay at the output of a gate using delays at each of its inputs based on a gain-based approach~\cite{sutherland1991logical}. While this abstraction simplifies the mapping process, it may result in suboptimal mapping quality~\cite{braytontechnology}. Recent works have applied reinforcement learning to technology mapping. For instance, MapTune~\cite{liu2024maptune} dynamically optimizes cell library selection in ASIC technology mapping using reinforcement learning, reducing the search space and improving mapping quality. Ye et al.~\cite{ye2024timing} propose a reinforcement learning framework that optimizes timing performance under error constraints to mitigate the effects of aging and variation. These approaches still rely on the fundamental mapping engines with load-independent models. Our DeepCell framework, by fusing information from pre- and post-mapping views, predicts actual loads, leading to more accurate delay predictions and improving the overall mapping quality.



%% file: 03_method.tex
\section{Methodology} \label{Sec:Method}
\subsection{Overview}
Fig.~\ref{fig:overview} presents the overview of DeepCell framework, which consists of a GNN-based AIG encoder, a GNN-based PM encoder and Transformer blocks. The framework operates in two stages to capture multiview information to enhance both AIG and PM netlist representations. In Stage 1, both the AIG and PM encoders are independently trained to extract AIG embeddings (\textit{AIG Emb.}) and PM netlist embeddings (\textit{PM Emb.}) from their respective input formats. In Stage 2, the embeddings from both views are concatenated into a token sequence, after which a random subset of tokens is masked. The Transformer blocks then reconstruct the masked tokens, effectively injecting multiview information into the node-level representations.

\subsection{PM Encoder}
Given a PM netlist, we convert it into graph format $\mathcal{G}^P=(\mathcal{V}^P, \mathcal{E}^P)$, where each standard cell is represented as node and each wire is treated as edge on the graph. 

\subsubsection{Node Features}
PM netlists consist of a wide variety of standard cells, making it impractical to represent them using one-hot encoding for each type of cell. Therefore, to address this, we embed the truth table of each standard cell into its corresponding node feature $x_{i}$. Formally, the node feature encoding is defined in Eq.~\eqref{eq:feature}, where $D$ is the dimension of node feature vector and $tt_{i}$ represents the 0/1 truth table vector of standard cell $i$. The truth table $tt_i$ is repeated until the node feature vector reaches the specified dimension $D$. In the default setting, we assign $D=64$, ensuring that this encoding mechanism is adaptable to PM netlists across various technology libraries and supports arbitrary logic units with up to 6 inputs, which is typical for most cells.
\begin{equation} \label{eq:feature}
    x_i = repeat(tt_i, D)
\end{equation}

For example, the cell \texttt{xor2\_1} defines XOR functionality with 2 inputs and 1 output. Its truth table, \texttt{0110}, is extracted from the technology library and expanded into a 64-dimensional feature vector by repeating the pattern. Thus, the node feature of the \texttt{xor2\_1} cell becomes \texttt{0110 0110 ... 0110} $\in \{0, 1\}^{64}$. This encoding mechanism can therefore handle arbitrary standard cells by representing their truth tables as binary vectors of a fixed length.

\subsubsection{Aggregator}
We introduce a DAG-based GNN to encode circuit graph into embedding vectors $\mathbf{H}^P$. For each cell $i\in \mathcal{V}^P$, its representation vector is denoted as $H_i^P = \{hs_i^P, hf_i^P\}$, where $hs_i^P$ and $hf_i^P$ are the structural and functional embeddings, $H_i^P \in \mathbf{H}^P$. 
To compute these embeddings, we propose two aggregators: $aggr^{s}$ for structural message aggregation and $aggr^{f}$ for functional message aggregation.

For structural embedding aggregation, $aggr^{s}$ is implemented using the GCN aggregator~\cite{kipf2016semi}, which aggregates messages from the predecessors of $i$. The initial structural embedding of each cell is randomly assigned and considered as its unique identifier. Here, $\mathcal{P}(i)$ denotes the set of fan-in cells of $i$: 
\begin{equation} \label{eq:aggrhs}
    hs_i = aggr^s(\{hs_j | j\in \mathcal{P}(i)\})
\end{equation}

For functional embedding aggregation, $aggr^{f}$ is implemented using a self-attention aggregator to distinguish the functionality of the predecessors. We discuss the effectiveness of different aggregators in Sec.~\ref{Sec:Exp:PMGNN}. Unlike AIG netlists, which consist solely of AND gates and inverters, PM netlists contain diverse standard cells. To account for this diversity, we differentiate cells using their node features $x_i$ and introduce an update function, $update$. Formally, the functional aggregation process in DeepCell is defined as Eq.~\eqref{eq:aggrhf}, with the $update$ function simply implemented as a multi-layer perceptron (MLP). 

\begin{equation} \label{eq:aggrhf}
    \begin{split}
        msg_i & = aggr^f(\{cat(hs_j, hf_j) | j \in \mathcal{P}(i)\}) \\ 
        hf_i & = update(msg_i, x_i) 
    \end{split}
\end{equation}

Finally, the embeddings of PM netlist are denoted as Eq.~\eqref{eq:pm}, where $\Phi^{P}$ is the above GNN-based PM encoder. 
\begin{equation} \label{eq:pm}
    \begin{split}
        \mathbf{H}^P & = \Phi^{P}(\mathcal{G}^P) \\ 
        H_i^P & \in \mathbf{H}^P, i \in \mathcal{V}^P 
    \end{split}
\end{equation}

\subsection{AIG Encoder}
Our DeepCell framework incorporates a multiview representation learning mechanism, allowing it to learn cell embeddings in PM netlists from an additional perspective provided by AIGs and refine AIG embeddings with complementary information from the PM netlists. 
As shown in Fig.~\ref{fig:overview}, given a PM netlist $\mathcal{G}^P=(\mathcal{V}^P, \mathcal{E}^P)$, we convert it into an AIG $\mathcal{G}^A=(\mathcal{V}^A, \mathcal{E}^A)$ using EDA tools. In our implementation, we use the ABC tool~\cite{brayton2010abc} with the command `\textit{strash}' to transform the PM netlist into an AIG through one-level structural hashing. We then utilize DeepGate2~\cite{shi2023deepgate2}, a well-trained and widely-used model for AIG representation learning, as the AIG encoder $\Phi^{A}$ to derive gate-level embeddings $H_j^A\in\mathbf{H}^A$, as defined in Eq.~\eqref{eq:aig}. Other AIG encoders, such as PolarGate~\cite{liu2024polargate}, DeepGate3~\cite{shi2024deepgate3} and HOGA~\cite{deng2024less}, are also compatible with this framework and are discussed in Section~\ref{Sec:Exp:AIGGNN}. In our default setting, we opt for DeepGate2~\cite{shi2023deepgate2} as the backbone AIG encoder due to the trade-off between model complexity and performance. 

\begin{equation} \label{eq:aig}
    \begin{split}
        \mathbf{H}^A & = \Phi^{A}(\mathcal{G}^A) \\
        H_j^A & \in \mathbf{H}^A, j \in \mathcal{V}^A 
    \end{split}
\end{equation}

\subsection{Mask Circuit Modeling}
\begin{figure} [!t]
    \centering
    \includegraphics[width=0.85\linewidth]{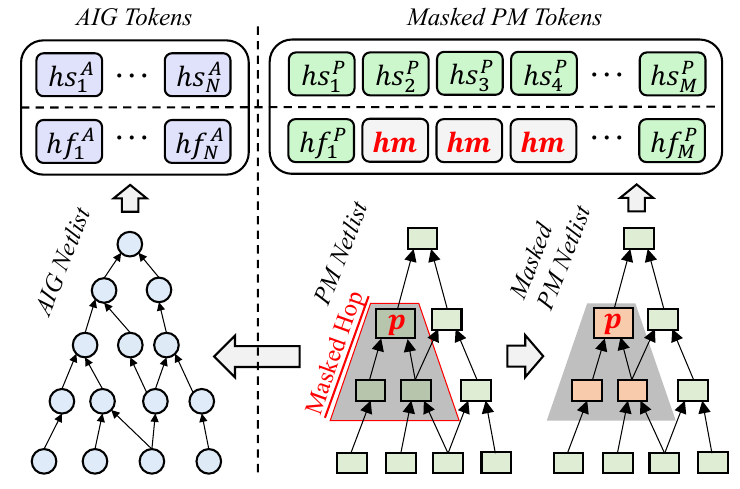}
    \caption{Mask circuit modeling of PM netlists}
    \label{fig:mcm}
    \vspace{-10pt}
\end{figure}

To incorporate multiview information into the GNN-based encoders, we propose the Masked Circuit Modeling (MCM) training mechanism. Unlike the conventional MLM in LLMs~\cite{devlin2018bert}, where certain tokens are randomly masked and then reconstructed using the remaining tokens, DeepCell masks circuit cones as illustrated in Fig.~\ref{fig:mcm}. To be specific, given a PM netlist $\mathcal{G}^P$ and the corresponding AIG $\mathcal{G}^A$, we randomly select node $p \in \mathcal{G}^P$ and extract a $k$-hop predecessors around node $p$, denoted as the Masked Hop $\mathcal{M}(p)$. All nodes $i \in \mathcal{M}(p)$ are considered as masked, and their functional embeddings $hf_i$ are replaced by the learnable masked token $hm$, while their structural embeddings $hs_i$ are preserving. 

As a result, we have $M+N$ tokens (see Eq.~\eqref{eq:token}) in total, where $M = |\mathcal{V}^P|$ and $N = |\mathcal{V}^A|$. All the selected masked nodes $p$ is in set $\mathcal{S}$. 

\begin{equation} \label{eq:token}
    \begin{split}
        {H_i^P} & = \{hs_i^P, hf_i^P\}, ~i\in \mathcal{V}^P, i \notin \mathcal{S} \\ 
        {H_i^P} & = \{hs_i^P, hm\}, ~i\in \mathcal{V}^P, i \in \mathcal{S} \\ 
        {H_j^A} & = \{hs_j^A, hf_j^A\}, ~j\in \mathcal{V}^A \\ 
    \end{split}
\end{equation}

Then, we process these $M+N$ tokens using a Transformer model $\mathcal{T}$. Formally, we define the input and output of the Transformer model as below. 
\begin{equation} \label{eq:transformer}
    \{\mathbf{H}^{A'}, \mathbf{H}^{P'}\} = \mathcal{T}(\{\mathbf{H}^A, \mathbf{H}^P\})
\end{equation}

Similarly, when the DeepCell framework is applied to refine the AIG embeddings, the MCM mechanism generates masked AIG tokens and non-masked PM tokens. It is important to note that recovering the node type is unnecessary, as it can be directly inferred by counting the fan-in degree in the AIG, where an AND gate always has 2 fan-in pins and a NOT gate has 1 fan-in pin. Consequently, the training objective is to reconstruct the functional embeddings using information from surrounding nodes and nodes in the other modality.

\subsection{Model pretaining}
To effectively learn rich multiview embeddings, we employ a two-stage training strategy. In the first stage, the GNN-based encoders are pretrained by predicting the logic-1 probability under random simulation, which provides rich supervision containing both structural and semantic information of Boolean logic \cite{li2022deepgate}. Specifically, we perform random simulation with $15,000$ patterns on PM netlist and record the logic-1 probability $prob_i, i \in \mathcal{V}^P$. Next, we use a simple 3-layer MLP to readout the embedding of each cell in PM netlist as the predicted probability. The training loss of PM encoder is defined in Eq.~\eqref{eq:stage1}. 
\begin{equation} \label{eq:stage1}
    \begin{split}
        \hat{prob^{P}_i} & = MLP(hf_i),~i \in \mathcal{V}^P\\ 
        L_{prob}^P & = L1Loss(prob_i, \hat{prob^{P}_i}) \\ 
    \end{split}
\end{equation}

We train the AIG encoder in a similar manner. The training loss of AIG encoder is defined in Eq.~\eqref{eq:stage1AIG}. 
\begin{equation} \label{eq:stage1AIG}
    \begin{split}
        \hat{prob^{A}_i} & = MLP(hf_i),~i \in \mathcal{V}^P\\ 
        L_{prob}^A & = L1Loss(prob_i, \hat{prob^{A}_i}) \\ 
    \end{split}
\end{equation}

In the second stage, we refine the cell embeddings by incorporating representations from the AIG view. We utilize the pretained and frozen AIG encoder to provide rich contextual information from the AIG. The training objective in this stage is to reconstruct the functionality of masked cells using information from their neighboring cells and the global AIG perspective. Accordingly, we define the MCM training loss in Eq.~\eqref{eq:stage2}, where the model is trained to recover the functional embeddings of the masked cells. 
\begin{equation} \label{eq:stage2}
    L_{mcm} = L1Loss(H_i^{P}, H_i^{P'})
\end{equation}

Alternatively, if DeepCell is employed to refine AIG embeddings and the PM encoder $\Phi^P$ is frozen. The loss function is modified to:
\begin{equation} \label{eq:stage2AIG}
    L_{mcm} = L1Loss(H_i^{A}, H_i^{A'})
\end{equation}

Consequently, the overall loss functions for both stages are defined as follows, where $w_{prob}$ and $w_{mcm}$ present the weights of these two training tasks. We assign $w_{prob}=1$ and $w_{mcm}=1$ in default setting. 
\begin{equation}
    \begin{split}
        L_{stage1} & = L_{prob}^P + L_{prob}^A \\
        L_{stage2} & = w_{prob} \cdot (L_{prob}^P + L_{prob}^A) + w_{mcm} \cdot L_{mcm} \\
    \end{split}
\end{equation}

%% file: 04_experiment.tex
\section{Experiments} \label{Sec:Experiment}
\subsection{Experiment Settings}
\subsubsection{Data Preparation} 
We employ ForgeEDA~\cite{shi2025forgeeda} and DeepCircuitX~\cite{li2025deepcircuitx} datasets, which collect open-source repository-level designs from GitHub. We randomly extract sub-circuits (both PM netlists and AIGs) containing up to 4,096 nodes for efficient model pre-training. It is important to note that DeepCell trained on small-scale sub-circuits can be generalized to larger circuits with acceptable runtime, as demonstrated in the practical downstream tasks discussed in Sec.~\ref{Sec:ECO} and Sec.~\ref{Sec:Mapping}. 

For model training and validation, we utilize three open-source technology libraries (\textit{Skywater130nm}~\cite{sky130nm}, \textit{GSCL45nm}~\cite{gscl45nm}, and \textit{GlobalFoundries(GF)180nm}~\cite{gf180nm}) to generate post-mapping netlists using commercial tool with a similar distribution (ID). Additionally, we prepare the testing dataset using another technology library, \textit{Nangate45nm}~\cite{nangate45nm}, which serves as an out-of-distribution (OOD) set. The statistics of our datasets are provided in TABLE~\ref{Tab:dataset}, where Dist. indicates whether the set has the same distribution as the training set.

\begin{table}[!t]
\caption{The statistics of DeepCell datasets} \label{Tab:dataset}
\centering
\begin{tabular}{@{}llll@{}}
\toprule
Dataset    & Dist. & Libraries             & \# Circuits \\ \midrule
Training      & ID          & \textit{Skywater130nm}, \textit{GSCL45nm}, \textit{GF180nm} & 16,072     \\
Validation & ID          & \textit{Skywater130nm}, \textit{GSCL45nm}, \textit{GF180nm} & 4,019      \\
Testing       & OOD         & \textit{Nangate45nm}               & 1,828      \\ \bottomrule
\end{tabular}
\vspace{-5pt}
\end{table}

\subsubsection{Evaluation Metrics} In the following experiments, we evaluate model performance in predicting the logic-1 probability under random simulation, a widely used metric for assessing circuit representation ability~\cite{li2022deepgate, liu2024polargate}. We calculate the average prediction error (PE) as Eq.~\eqref{eq:pe}. 
\begin{equation} \label{eq:pe}
    PE = \frac{1}{|\mathcal{V}^P|}\sum_{i\in\mathcal{V}^P}|prob_i - \hat{prob}_{i}|
\end{equation}

Additionally, we assess model performance on mask circuit modeling tasks, where the average reconstruction error (RE) is defined as Eq.~\eqref{eq:re}, where all the select nodes $p$ are in set $\mathcal{S}$ and the masked hop is $\mathcal{M}(p)$. Lower values ($\downarrow$) indicate better performance for both metrics. 
\begin{equation} \label{eq:re}
    RE = \frac{1}{\sum|\mathcal{M}(p)|}\sum_{i\in \mathcal{M}(p)}|hf_i^P-hf_i^{P'}|, p\in\mathcal{S} 
\end{equation}

\subsubsection{Model Implementation and Training}
The Transformer-based model used to refine cell embeddings consists of 4 Transformer blocks, each with 8 attention heads. After encoding, each cell is represented by a 128-dimensional structural embedding and a 128-dimensional functional embedding. DeepCell is pretrained for 60 epochs in Stage 1 and an additional 60 epochs in Stage 2 to ensure convergence. The pretraining process is conducted with a batch size of 128 using 8 Nvidia A800 GPUs. We employ the Adam optimizer with a learning rate of $10^{-4}$. 

\subsection{Experimental Results}

\subsubsection{Hyperparameter Analysis}
We investigate the optimal settings for model pretraining by exploring two key hyperparameters in MCM: the number of selected nodes to be masked, $|\mathcal{S}|$, and the size of the masked hop, $k$. To evaluate performance, we pretrain the model using various hyperparameter combinations, where $\mathcal{|S|}=\theta \cdot |\mathcal{V}^P|$ and $\theta = 1\%, 5\%, 10\%, 20\%$ of total nodes in PM netlist, with hop sizes of $k=4$ or $k=6$. 

\begin{table}[!t]
\caption{Effect of Different Hyperparameter} \label{TAB:mask}
\centering
\begin{tabular}{@{}l|ll|ll@{}}
\toprule
\multicolumn{1}{c|}{\multirow{2}{*}{$\theta$}} & \multicolumn{2}{c|}{$k=4$}            & \multicolumn{2}{c}{$k=6$} \\
\multicolumn{1}{c|}{}                                                          & PE              & RE              & PE        & RE        \\ \midrule
0.01                                                                           & \textbf{0.0322} & 0.0099          & \textbf{0.0380}    & \textbf{0.0211}    \\
0.05                                                                           & 0.0323          & \textbf{0.0097} & 0.0418    & 0.0257    \\
0.10                                                                           & 0.0334          & 0.0110          & 0.0552    & 0.0794    \\
0.20                                                                           & 0.0446          & 0.0398          & 0.0682    & 0.1035    \\ \bottomrule
\end{tabular}
\vspace{-5pt}
\end{table}

TABLE~\ref{TAB:mask} presents the results for different values of $\theta$ and $k$. First, the reconstruction error (RE) increases with a larger masking hop size. For example, when $\theta =0.05$, the RE for $k=6$ is 0.0257, which is 164.95\% higher than that for $k=4$ (RE=0.0097). Second, masking a smaller number of nodes consistently reduces both the RE and PE. However, using a smaller $\theta$ makes the task less challenging and diminishes its effectiveness as a pretraining objective. Based on these observation, we select $\theta=0.05$ and $k=4$ as a trade-off between task complexity and model performance in the following experiments. 

\subsubsection{Effect of Multiview Learning on PM Netlist} \label{Sec:Exp:PMGNN}
We investigate the impact of incorporating the multiview information on PM netlist representation learning. Specifically, we compare the full multiview MCM training strategy (w/ multiview) with a baseline that uses only the PM encoder without refining embeddings through multiview training (w/o multiview). The RE metric is not available in the w/o multiview setting. 
To the best of our knowledge, no prior work has focused on learning general-purpose representations of post-mapping (PM) netlists composed of standard cells. To evaluate the effectiveness and generalization ability of our proposed Mask Circuit Modeling (MCM) training strategy, we incorporate different GNN-based aggregators: Convolutional Sum (Conv. Sum)~\cite{selsam2018learning}, Attention~\cite{velivckovic2017graph} and the aggregator in DeepGate2 (DG2)~\cite{shi2023deepgate2}, into PM encoders within the DeepCell framework. 

\begin{table}[!t]
\caption{Effect of Multiview Learning on PM Netlist Encoder} \label{TAB:PM}
\centering
\begin{tabular}{@{}llllll@{}}
\toprule
Aggr      & Multview & Dist. & PE              & Red.  & RE     \\ \midrule
Conv. Sum & w/o      & ID    & 0.0344          & \--      & \--      \\
          & w/       & ID    & 0.0312          & 9.41\% & 0.0081 \\
          & w/o      & OOD   & 0.0380          & \--      & \--      \\
          & w/       & OOD   & 0.0360          & 5.29\% & 0.0096 \\ \midrule
Attention & w/o      & ID    & 0.0337          & \--      & \--      \\
          & w/       & ID    & 0.0318          & 5.65\% & 0.0076 \\
          & w/o      & OOD   & 0.0350          & \--      & \--      \\
          & w/       & OOD   & 0.0325          & 7.04\% & 0.0087 \\ \midrule
DG2       & w/o      & ID    & 0.0314          & \--      & \--      \\
          & w/       & ID    & \textbf{0.0286} & 8.92\% & 0.0082 \\
          & w/o      & OOD   & 0.0316          & \--      & \--      \\
          & w/       & OOD   & 0.0295          & 6.68\% & 0.0089 \\ \bottomrule
\end{tabular}
\vspace{-10pt}
\end{table}

TABLE~\ref{TAB:PM} compares the performance of the models with (w/) and without (w/o) multiview training. The column `Red.' shows the PE reduction achieved by incorporating multiview information. First, compared to other PM encoders, whether using multiview training or not, the DG2 aggregator outperforms alternatives with lowest PE of 0.0286. We employ such aggregator in the following applications. Second, all models with multiview training outperform their counterparts without multiview training in terms of PE. For example, Conv. Sum w/ multiview reduces PE by 9.41\%, from 0.0344 to 0.0312. Third, our PM netlist encoder performs similarly on both ID and OOD datasets, highlighting the generalization ability of DeepCell across different technology libraries.

\subsubsection{Effect of Multiview Learning on AIG} \label{Sec:Exp:AIGGNN}
We further explore the impact of multiview learning on enhancing AIG representations. Specifically, we compare the performance of AIG encoders with (w/) and without (w/o) multiview training using the PE metric. Four representative AIG encoders—PolarGate~\cite{liu2024polargate}, HOGA~\cite{deng2024less}, DeepGate2~\cite{shi2023deepgate2}, and DeepGate3~\cite{shi2024deepgate3}—along with a general GCN model~\cite{kipf2016semi}, are evaluated in the following experiments. As shown in TABLE~\ref{TAB:AIG}, incorporating multiview training significantly improves model performance. For example, DeepGate2 achieves a PE of 0.0346 without multiview training, but with multiview training, the PE is reduced by 28.03\%, reaching 0.0249. This demonstrates the considerable benefit of adding information from the PM view. 

Furthermore, since the outputs of the GNN-based AIG encoder are treated as AIG embeddings within the DeepCell framework (see Fig.~\ref{fig:overview}), the runtime and memory usage for AIG representation learning during inference remain identical to those of the baseline models. This ensures that the introduction of multiview learning does not incur any additional computational overhead, while still delivering substantial performance improvements. 

\begin{table}[!t]
\caption{Effect of Multiview Learning on AIG Encoder} \label{TAB:AIG}
\centering
\begin{tabular}{@{}lllll@{}}
\toprule
Model     & Multview & PE              & Red.   & RE     \\ \midrule
GCN~\cite{kipf2016semi}       & w/o      & 0.0599          & \--       & \--       \\
                              & w/       & 0.0540          & 9.85\%  & 0.0081 \\ \midrule
PolarGate~\cite{liu2024polargate} & w/o      & 0.0318          & \--       & \--       \\
                                  & w/       & 0.0309          & 2.83\%  & 0.0076 \\ \midrule
HOGA~\cite{deng2024less}      & w/o      &  0.0704          & \--       &  \--      \\
                              & w/       &  0.0685          & 2.70\%       & 0.0084       \\ \midrule
DeepGate2~\cite{shi2023deepgate2} & w/o      & 0.0346          & \--       & \--       \\
                                  & w/       & \underline{0.0249}    & 28.03\% & 0.0087 \\ \midrule
DeepGate3~\cite{shi2022deeptpi} & w/o      & 0.0263          & \--       & \--       \\
                                & w/       & \textbf{0.0244} & 7.22\%  & 0.0093 \\ \bottomrule
\end{tabular}
\vspace{-10pt}
\end{table}

%% file: 041_ECO.tex
\section{Downstream Task: Functional ECO} \label{Sec:ECO}

\subsection{Preliminary}
Functional Engineering Change Orders (ECO) is an essential task in EDA, where an existing circuit design is updated to meet new target specifications without requiring a full redesign. Specifically, given an \textit{original} PM netlist $\widetilde{\mathcal{G}}^P$ and the target patch location, functional ECO focuses on adding a patch to construct another netlist $\mathcal{G}^P$, which meets the functionality of the \textit{golden} design. The method introduced in \cite{dao2018efficient} for functional ECO wins the ICCAD contest in this domain. The algorithm first identifies the primary outputs (POs) that are reachable from the target patch location. For each reachable PO, the method then calculates the complete driven cones, identifying the candidate signals that could potentially be involved in the patch generation. Finally, the selected candidate signals are used to prove whether there is a solution using SAT-based approach~\cite{wu2010robust}.

However, the computational cost of generating the patch and verifying the solution using SAT solvers becomes prohibitively expensive as the number of candidate signals increases. Although \cite{dao2018efficient} restricts the SAT-based approach to generate the patch using only a fixed number of low-cost candidate signals in order to save runtime, this constrained set often leads to situations where certain ECO problems cannot be solved. The challenge, therefore, lies in efficiently reducing the number of candidate signals considered, while still ensuring that the ECO problem can be solved within acceptable computational constraints. 

\subsection{Model Finetuning}
DeepCell is employed to effectively narrow down the candidate signal space by learning which signals are most likely to be relevant for the patch generation task. Intuitively, the original circuit that requires patch insertion in ECO task is treated as PM netlist with masking, while the golden circuit provide the complete view. Therefore, we finetune DeepCell to predict the driven signals of the patch with the view from both original PM netlist and golden AIG. 

\begin{figure}
    \centering
    \includegraphics[width=0.90\linewidth]{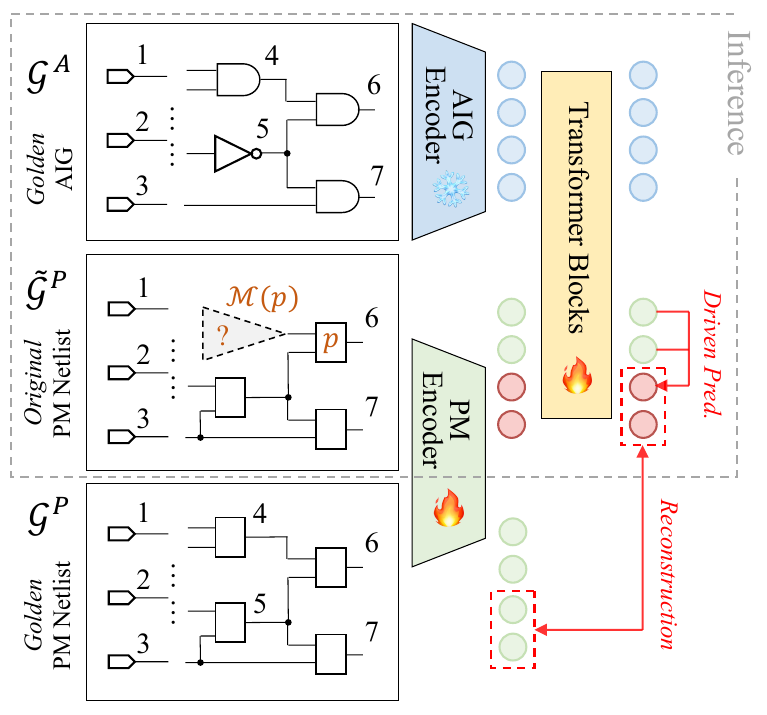}
    \caption{DeepCell finetuning for functional ECO}
    \label{fig:eco}
    \vspace{-5pt}
\end{figure}

As shown in Fig.~\ref{fig:eco}, given a PM netlist $\mathcal{G}^P$, we randomly select a cell $p$ within the netlist and remove its entire driven cone from this node to PI. The removed area, $\mathcal{M}(p)$, is treated as the ground-truth path in ECO. Consequently, the patch-removed netlist is denoted as $\widetilde{\mathcal{G}}^P$, while the corresponding AIG for the PM netlist $\mathcal{G}^P$ is denoted as $\mathcal{G}^A$. Next, we use the PM encoder $\Phi^P$ to encode both $\widetilde{\mathcal{G}}^P$ and $\mathcal{G}^P$, obtaining the corresponding embeddings $\widetilde{\mathbf{H}}^{P}$ and $\mathbf{H}^P$. The AIG encode $\Phi^A$ is employed to encode the AIG netlist $\mathcal{G}^A$, where the AIG embeddings $\mathbf{H}^A$ serves as the view from golden implementation. The embeddings of the patch-removed PM netlist are then refined using the AIG embeddings using Transformer Blocks to produce the refined tokens (see Eq.~\eqref{eq:eco}). 
\begin{equation} \label{eq:eco}
    \begin{split}
        \{\mathbf{H}^{A'}, \widetilde{\mathbf{H}}^{P'}\} & = \mathcal{T}(\{\mathbf{H}^{A}, \widetilde{\mathbf{H}}^{P}\})
    \end{split}
\end{equation}

The finetuning tasks are defined as patch reconstruction (\textit{Reconstruction}) and driven signal prediction (\textit{Driven Pred.}). In the first task, we treat the embeddings of the golden PM netlist $\mathcal{G}^P$ as the ground truth and define the reconstruction loss $L_{rec}$. The goal is to minimize the difference between the reconstructed patch-removed PM netlist and the original golden PM netlist. In the second task, we focus on driven signal prediction. Here, the refined embeddings $\widetilde{\mathbf{H}}^{P'}$ are used to predict whether a certain cell in the netlist is driven by the cell $p$ at the patch insertion location. 

\begin{equation} \label{eq:pr}
    \begin{split}
        L_{rec} & = L1Loss(H_i^{P'}, \widetilde{H}_i^{P'}),~i\in\mathcal{M}(p) \\
    L_{drv} & = BCE(MLP(\{\widetilde{H}_i^{P'}, \widetilde{H}_p^{P'}\}), 0/1), i\in\mathcal{V}^P
    \end{split}
\end{equation}

We finetune the pretrained DeepCell using 8,000 circuits only over 10 epochs (about 2 GPU hours) with loss function $L=L_{rec} + L_{drv}$. 
To ensure efficiency during inference while accepting a slight trade-off in accuracy, we implement the Transformer Blocks using Linear Transformer~\cite{kitaev2020reformer}. During inference, we only process the original PM netlist $\widetilde{\mathcal{G}}^P$ (the patch-removed netlist used during finetuning) and the golden AIG $\mathcal{G}^A$. Our model produces a list of floating-point numbers, each representing the probability that a given cell should be included in the candidate signal set.

\subsection{Experiment Settings}
Our model is equipped into ABC~\cite{brayton2010abc} as a plug-in and integrated into the command \textit{`runeco'}, which is an efficient SAT-based solution that won the first place in the 2017 ICCAD contest problem A~\cite{huang20172017} and added the SAT-based exact pruning method~\cite{dao2018efficient}. To the best of our knowledge, this is the latest known open-source approach for functional ECO. Our method only predicts driving relations to prune candidate signals, make it compatible with the other ECO engines~\cite{kravets2020learning, liu2023feep}. As a general plug-in, DeepCell can be seamlessly integrated into these frameworks if source code is accessible. 

In the following experiments, DeepCell demonstrates its effectiveness by selecting 1,000 driving nodes based on predicted probabilities as candidate signals. The benchmarks are using ISCAS-85/89~\cite{brglez1989combinational}, ITC-99~\cite{davidson1999characteristics}, IWLS-2005~\cite{albrecht2005iwls}, OpenCore and LGSynth-93 benchmarks and real-world ECO problems from industrial design. All experiments including model inference are performed on a single core of 2.10 GHz Intel(R) Core(TM) i7-14700F CPU with 32GB memory. 


\subsection{Main Results}
As shown in TABLE~\ref{compare on ICCAD 2017 contest}, the columns labeled \text{`gate(F)'} and \text{`gate(G)'} correspond to the number of gates in the original circuit and the golden circuit, respectively. The \text{`w/o DeepCell'} and \text{`w/ DeepCell'} sections lists the effects of \textit{`runeco'} before and after inserting DeepCell. The ABC tool runtime is labeled as \text{time(s)}, the model runtime is labeled as \text{model(s)}, and the \text{overall(s)} column accumulates both. The column \text{target} represents the number of patch insert locations. 
We limit the maximum running time to two hours. These timeout units are labeled as `-'.

Overall, DeepCell reduces the average cost of patches by 2.77\% and the average number of gates by 11.60\%. Additionally, the required runtime decreases by 25.63\%, with no units experiencing increased costs or gate counts. Notably, in the top-performing unit 16, our model achieves a remarkable 27\% reduction in cost and a 40\% reduction in gate count. These results are attributed to the ability of effectively capturing node relationships, enabling it to calculate the probability of each node for every target in multi-target units. 

For small and easy cases (unit 1-4), incorporating DeepCell may slightly increase the runtime due to the high proportion of model inference time. However, as the circuit size increases, the number of potential candidate nodes can be significantly reduced by utilizing the probabilities provided by DeepCell, which in turn accelerates the process. As a result, out of 20 units, the running time increased in 10 instances. The most efficient application (unit 5) achieved a 70\% reduction in runtime, while the least efficient case only saw an increase of 0.52 seconds (unit 20).

\begin{table}[!t]
  \centering
  \caption{Performance Comparison between w/ DeepCell and w/o DeepCell on Find Feasible ECO Solution}
    \tabcolsep=0.014\linewidth
    \begin{tabular*}{\linewidth}{ccccccc}
    \toprule
    \multirow{2}[2]{*}{Circuit name} & \multicolumn{3}{c}{\text{w/o DeepCell}} & \multicolumn{3}{c}{\text{w/ DeepCell}} \\
          & cost  & support size & time(s) & cost  & support size & time(s) \\
    \midrule
    unit 6 & \--     & \--     & $>$12600 & 2500  & 42    & 538.06 \\
    unit 10 & 63    & 16    & 48.64 & 63    & 16    & 29.2 \\
    unit 11 & 54    & 3     & 3977.58 & 36    & 2     & 2379.78 \\
    unit 19 & \--     & \--     & $>$12600 & \--     & \--     & $>$12600 \\
    \bottomrule
    \end{tabular*}%
  \label{compare in large unit}%
  \vspace{-5pt}
\end{table}%


For four units encountering timeout (unit 6, 10, 11, 19), we conduct further analysis in TABLE~\ref{compare in large unit}. The results regarding the cost, support size (number of candidate signals used), and runtime for the first single patch insertion of these units are collected. We increase the limit time to three and a half hours (12,600s). Except for unit 19, the other three units are able to provide a viable solution more quickly, and the quality of these solutions are superior. To be specific, unit 6 is timeout on w/o DeepCell ECO engine, but can be efficiently solved in 538.06s with DeepCell. 

\input{table/eco}
\input{table/map}

%% file: table/eco.tex
\begin{table*}[!t]
\centering
  \caption{Functional ECO Performance Comparison between w/ DeepCell and w/o DeepCell on ICCAD’17 Contest Benchmarks} \label{compare on ICCAD 2017 contest}
    \tabcolsep=0.01\linewidth
\begin{tabular}{@{}cccccc|ccc|ccccc@{}}
\toprule
\multirow{2}{*}{Circuit name} & \multicolumn{5}{c|}{Circuit information}                                                                          & \multicolumn{3}{c|}{w/o DeepCell} & \multicolumn{5}{c}{w/ DeepCell}                      \\
                              & PI                   & PO                   & gate(F)              & gate(G)              & tatget                & cost      & gate    & time(s)    & cost   & gate   & model(s) & time(s) & overall(s) \\ \midrule
unit 1                        & 3                    & 2                    & 6                    & 6                    & 1                     & 4         & 1       & 0.37        & 4      & 1      & 0.06      & 0.37     & 0.43        \\
unit 2                        & 157                  & 64                   & 1,120                & 1,219                & 1                     & 17        & 4       & 0.92        & 17     & 4      & 0.41      & 1.01     & 1.42        \\
unit 3                        & 411                  & 128                  & 2,074                & 1,929                & 1                     & 80        & 3       & 0.46        & 80     & 3      & 0.38      & 0.40     & 0.78        \\
unit 4                        & 11                   & 6                    & 75                   & 77                   & 1                     & 32        & 5       & 0.49        & 32     & 2      & 0.10      & 0.55     & 0.65        \\
unit 5                        & 450                  & 282                  & 24,357               & 21,056               & 2                     & 47        & 30      & 46.30        & 47     & 29     & 7.36      & 15.39    & 22.75       \\
unit 6                        & 99                   & 128                  & 13,828               & 11,812               & 2                     & \--         & \--       & \--           & \--      & \--      & 5.13      & \--        & \--           \\
unit 7                        & 207                  & 24                   & 2,944                & 1,721                & 1                     & 284       & 2       & 8.50         & 284    & 2      & 0.47      & 6.88     & 7.35        \\
unit 8                        & 179                  & 64                   & 2,513                & 3,337                & 1                     & 78        & 4       & 4.10         & 78     & 3      & 0.63      & 3.17     & 3.80        \\
unit 9                        & 256                  & 245                  & 5,849                & 4,657                & 4                     & 50        & 35      & 1.33        & 50     & 26     & 0.65      & 1.70     & 2.35        \\
unit 10                       & 32                   & 129                  & 1,581                & 1,956                & 2                     & \--         & \--       & \--           & \--      & \--      & 0.73      & \--        & \--           \\
unit 11                       & 48                   & 50                   & 2,057                & 2,160                & 8                     & \--         & \--       & \--           & 2,312  & 746    & 0.71      & 6,975.30 & 6,976.01    \\
unit 12                       & 46                   & 27                   & 13,804               & 821                  & 1                     & 104       & 1       & 1.02        & 104    & 1      & 0.36      & 0.71     & 1.07        \\
unit 13                       & 25                   & 39                   & 369                  & 426                  & 1                     & 3,467     & 9       & 1.50         & 3,467  & 9      & 0.12      & 1.33     & 1.45        \\
unit 14                       & 17                   & 15                   & 1,981                & 1,006                & 12                    & 95        & 41      & 5.43        & 95     & 41     & 0.23      & 6.72     & 6.95        \\
unit 15                       & 198                  & 14                   & 1,886                & 2,262                & 1                     & 191       & 11      & 1.93        & 191    & 11     & 0.37      & 0.73     & 1.10        \\
unit 16                       & 417                  & 214                  & 2,371                & 9,324                & 2                     & 278       & 15      & 16.58       & 204    & 9      & 1.98      & 14.67    & 16.65       \\
unit 17                       & 136                  & 31                   & 2,910                & 2,052                & 8                     & 434       & 79      & 5.04        & 434    & 63     & 0.43      & 6.01     & 6.44        \\
unit 18                       & 245                  & 100                  & 4,860                & 3,881                & 1                     & 18        & 1       & 5.14        & 18     & 1      & 0.97      & 2.44     & 3.41        \\
unit 19                       & 99                   & 128                  & 13,349               & 10,787               & 4                     & \--         & \--       & \--           & \--      & \--      & 4.02      & \--        & \--           \\
unit 20                       & 1,874                & 7,105                & 30,876               & 34,002               & 4                     & 136       & 6       & 0.60         & 120    & 5      & 7.80      & 1.12     & 8.92        \\ \midrule
Geomean$^*$                       & \multicolumn{1}{l}{} & \multicolumn{1}{l}{} & \multicolumn{1}{l}{} & \multicolumn{1}{l}{} & \multicolumn{1}{l|}{} & 147.88    & 9.21    & 4.49        & 143.78 & 8.14   & 0.61      & 3.34     & 4.16        \\
Reduction                     & \multicolumn{1}{l}{} & \multicolumn{1}{l}{} & \multicolumn{1}{l}{} & \multicolumn{1}{l}{} & \multicolumn{1}{l|}{} & \--          & \--        & \--            & \textbf{2.77\%}  & \textbf{11.60\%} & \--          & \textbf{25.63\%}   & \textbf{7.23\%}       \\ \bottomrule
\end{tabular}
\begin{tablenotes}
\footnotesize
\item[1] {$^*$} Exclude the easy cases w/o DeepCell time less than 1s and the hard cases cannot be addressed. 
\end{tablenotes}
\end{table*}


%% file: table/map.tex
\begin{table*}[!ht]
    \centering
    \caption{Technology Mapping Performance Comparison between w/ DeepCell and w/o DeepCell on EPFL Benchmark Suite}
    \tabcolsep = 0.007\linewidth
\begin{tabular}{@{}lccc|rrrr|rrrr@{}}
\toprule
\multirow{3}{*}{Circuit} & \multirow{3}{*}{PI/PO} & \multicolumn{1}{l}{} & \multicolumn{1}{l|}{} & \multicolumn{4}{c|}{\textit{Skywater130nm}}                                                                                                                               & \multicolumn{4}{c}{\textit{Nangate45nm}}                                                                                                                    \\ \cmidrule(l){5-12} 
                           &                        & \multicolumn{2}{c|}{Origin AIG}              & \multicolumn{2}{c}{\text{w/o DeepCell}}                                      & \multicolumn{2}{c|}{\text{w/ DeepCell}}                                       & \multicolumn{2}{c}{\text{w/o DeepCell}}                         & \multicolumn{2}{c}{\text{w/ DeepCell}}                                       \\
                           &                        & nodes                & levels                & \text{area(nm$^2$)}   & \text{delay(ps)}  & \text{area(nm$^2$)}  & \text{delay(ps)}     & \text{area(nm$^2$)}      & delay(ps)            & \text{area(nm$^2$)} & \text{delay(ps)}    \\ \midrule
adder                                          & 256/129                & 1,310                & 97                   & 4,790.84                 & 7,314.47                  & 4,900.95                 & 7,362.27                  & 966.38                   & 1,311.01                  & \textbf{964.78}          & 1,577.49                  \\
arbiter                                        & 256/129                & 4,589                & 14                   & 19,991.67                & 1,641.73                  & \textbf{16,934.99}       & 1,930.43                  & 3,777.73                 & 446.96                    & \textbf{3,448.16}        & 481.82                    \\
bar                                            & 135/128                & 3,600                & 11                   & 9,972.06                 & 3,848.12                  & \textbf{9,658.01}        & \textbf{3,658.24}         & 2,328.56                 & 2,011.48                  & \textbf{1,810.66}        & \textbf{1,974.39}         \\
cavlc                                          & 10/11                  & 600                  & 10                   & 1,870.54                 & 1,313.52                  & 1,875.55                 & \textbf{1,164.15}         & 437.04                   & 337.64                    & \textbf{391.82}          & \textbf{318.92}           \\
ctrl                                           & 7/26                   & 85                   & 6                    & 332.82                   & 827.57                    & 357.84                   & \textbf{602.28}           & 64.37                    & 199.91                    & 67.83                    & \textbf{155.97}           \\
dec                                            & 8/256                  & 304                  & 3                    & 1,203.65                 & 642.70                    & 1,203.65                 & \textbf{643.65}           & 252.97                   & 194.92                    & \textbf{247.91}          & \textbf{152.91}           \\
div                                            & 128/128                & 58,798               & 1,621                & 227,563.25               & 211,175.92                & 208,626.33               & 242,915.17                & 45,573.25                & 50,162.67                 & \textbf{41,685.66}       & 52,871.67                 \\
i2c                                            & 147/142                & 1,013                & 8                    & 3,624.73                 & 1,672.64                  & \textbf{3,434.54}        & \textbf{1,547.64}         & 653.30                   & 433.73                    & 676.97                   & \textbf{392.02}           \\
int2float                                      & 11/7                   & 208                  & 9                    & 758.23                   & 731.89                    & \textbf{670.64}          & \textbf{730.83}           & 151.09                   & 180.92                    & \textbf{137.52}          & 209.27                    \\
log2                                           & 32/32                  & 38,817               & 202                  & 159,729.44               & 43,278.46                 & \textbf{123,851.28}      & \textbf{40,329.39}        & 31,323.36                & 9,472.56                  & \textbf{25,398.21}       & \textbf{8,971.46}         \\
max                                            & 512/130                & 4,378                & 29                   & 16,502.08                & 25,513.67                 & \textbf{13,311.52}       & \textbf{19,598.67}        & 3,498.96                 & 5,569.14                  & \textbf{2,675.96}        & \textbf{3,294.48}         \\
mem\_ctrl                                       & 1204/1231              & 33,661               & 34                   & 109,762.77               & 12,184.23                 & \textbf{106,116.77}      & 13,581.43                 & 22,063.90                & 4,281.56                  & \textbf{21,394.11}       & 4,534.66                  \\
multiplier                                     & 128/128                & 31,817               & 125                  & 135,399.86               & 17,563.55                 & \textbf{102,020.34}      & 18,984.09                 & 26,249.41                & 4,059.93                  & \textbf{20,924.09}       & 4,120.63                  \\
priority                                       & 128/8                  & 457                  & 11                   & 1,527.72                 & 1,813.81                  & 1,569.00                 & \textbf{1,199.12}         & 337.29                   & 300.68                    & \textbf{323.46}          & 360.82                    \\
router                                         & 60/30                  & 159                  & 12                   & 469.20                   & 1,385.41                  & 505.48                   & \textbf{1,094.54}         & 96.56                    & 354.57                    & 101.61                   & \textbf{274.00}           \\
sin                                            & 24/25                  & 6,896                & 100                  & 32,406.08                & 20,786.12                 & \textbf{23,842.87}       & \textbf{18,327.04}        & 6,750.55                 & 4,593.38                  & \textbf{4,831.36}        & \textbf{3,775.06}         \\
sqrt                                           & 128/64                 & 30,670               & 1,637                & 119,448.30               & 406,476.56                & 105,303.49               & 420,115.69                & 24,054.11                & 95,101.24                 & \textbf{21,038.21}       & \textbf{91,700.94}        \\
square                                         & 64/128                 & 17,405               & 73                   & 74,765.45                & 7,488.54                  & \textbf{66,362.40}       & \textbf{6,660.44}         & 12,773.05                & 2,118.08                  & 12,783.16                & \textbf{1,729.88}         \\
voter                                          & 1001/1                 & 9,375                & 48                   & 52,293.90                & 5,474.38                  & \textbf{40,252.36}       & 5,677.99                  & 10,948.29                & 1,230.04                  & \textbf{7,580.47}        & 1,287.86                  \\ \midrule
\multicolumn{1}{l}{Geomean}           &                        &                      &                       & 11,676.95                             & 6,336.12                               & \text{10,586.68}                    & \text{5,939.16}                       & 2,366.03                              & 1591.05                  & \text{2115.02}                     & \text{1495.16}                      \\
\multicolumn{2}{l}{Area Delay Product (ADP) }   & \multicolumn{1}{l}{} & \multicolumn{1}{l|}{} &                                       & 73,986,525.76                          &                                       & \text{62,875,983.61}                  &                                       & 3,764,465.11               &                                       & \text{3,162,290.41}                  \\
\multicolumn{1}{l}{Reduction}             & \multicolumn{1}{l}{}   & \multicolumn{1}{l}{} & \multicolumn{1}{l|}{} &                                       & -                                      &                                       & \textbf{15.02\%}                           &                                       & -                         &                                       & \textbf{16.00\%}                          \\ \bottomrule
\end{tabular}
\label{mapping}
\end{table*}

%% file: 042_Mapping.tex
\section{Downstream Task: Technology Mapping} \label{Sec:Mapping}

\subsection{Preliminary}
Technology mapping is a crucial step in chip design flow, where an abstract representation of a circuit (typically in AIG) is mapped onto a specific technology library, consisting of standard cells. The goal of technology mapping is to find an optimal realization of the circuit using the available library cells, minimizing cost metrics such as area, delay, and power consumption, while ensuring that the functionality of the original design is preserved. 

Cut-based technology mapping~\cite{hinsberger1998boolean} with delay estimation is one of the most common approaches for achieving optimized results through dynamic programming~\cite{kukimoto1998delay}. In this process, nodes in the AIG are processed in topological order, with the objective of finding a delay-optimal match for each node in the target technology library. The delay values used to evaluate the quality of these matches are derived from a load-independent delay model, which assigns fixed delay values to gates based on their input delays and gain-based approximations. However, this approach often results in suboptimal outcomes because the delay estimation is inaccurate due to the absence of the actual capacitive load of the gates. Unfortunately, the load cannot be fully determined during the mapping process, as the detailed fan-out connections are not yet available. Consequently, predicting the fan-out connections to estimate capacitive load has the potential to enhance technology mapping.


\subsection{Model Finetuning}
\begin{figure}
    \centering
    \includegraphics[width=0.65\linewidth]{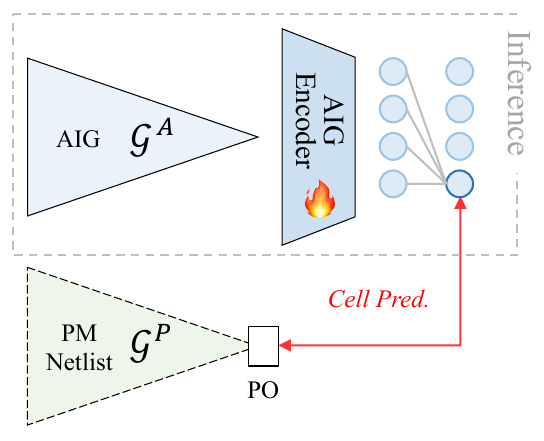}
    \caption{DeepCell finetuning for technology mapping}
    \label{fig:mapping}
  \vspace{-5pt}
\end{figure}

Since DeepCell captures the correlation between PM netlist and AIGs, we finetune the AIG encoder within DeepCell framework to predict the cell type implementing PO of the pre-mapping AIG cone. 
As illustrated in Fig.~\ref{fig:mapping}, given the AIG $\mathcal{G}^A$ and its corresponding PM netlist, we first use the AIG encoder to encode $\mathcal{G}^A$ into node-level embedding vectors $\mathbf{H}^A$. These embedding vectors are then processed using a pooling function, and an MLP (multi-layer perceptron) is applied to predict the cell type of the PO. The cell type prediction loss $L_{ctp}$ is defined in Eq.~\eqref{eq:ct}, where $y'$ represents prediction and $y$ denotes the ground truth cell type of the PO in the PM netlist.

To prepare the training samples, we randomly select 8,000 cells from various PM netlists and record their corresponding cell types as labels. For each selected cell, the corresponding cone from the PIs to the cell is extracted and mapped into the AIG format. The pretrained AIG encoder is then fine-tuned for 60 epochs over 0.5 hours using 8 Nvidia A800 GPUs (about 4 GPU hours). To further investigate the flexibility and versatility of DeepCell, we finetune it for two different mapping flows, utilizing two different technology libraries: \textit{Skywater130nm} (which was part of the pretraining) and \textit{Nangate45nm} (an out-of-distribution library). After finetuning, both models achieve cell type prediction accuracy of over 90\%. 

\begin{equation} \label{eq:ct}
    \begin{split}
        \mathbf{H}^A & = \Phi^A(\mathcal{G}^A) \\
         y' & = MLP(\text{Pooling}(\mathbf{H}^A)) \\
         L_{ctp} & = BCE(y', y)
    \end{split}
\end{equation}


\subsection{Experiment Settings}
Our DeepCell framework is used to enhance the \textit{`\&nf'} command in ABC~\cite{brayton2010abc}, which is currently recognized as the state-of-the-art open-source technology mapping implementations. DeepCell predicts the cell type connectivity for each target node to generate more accurate delay estimation. We validate our approach using the EPFL combinational benchmark suite~\cite{amaru2015epfl}. The input AIGs undergo preprocessing through the following optimized synthesis flow prior to technology mapping:\:\textit{`resyn2rs; compress2rs; \&sopb'}, iterating 6 times. We exclude the benchmark circuit `hyp' since a known bug in ABC~\cite{calvino2024practical}. Our experiments are conducted on two open-source libraries: \textit{Skywater130nm} and \textit{Nangate45nm}, load into ABC through the command \textit{`read\_lib -G 250'}.


The finetuned DeepCell is capable of delivering predictions for nearly all benchmark circuits within 1 second. As technology mapping itself is a fast and one-time effort, temporal comparisons are of little consequence.
The experimental environment maintains the same configurations as Sec.~\ref{Sec:ECO}.

\subsection{Main Results}
As shown in TABLE~\ref{mapping}, the technology mapping engine with DeepCell (w/ DeepCell) outperforms the baseline (w/o DeepCell). 
First, by predicting the actual load for more accurate delay estimation, DeepCell improves delay performance compared to the baseline engine. For example, 11 out of 19 benchmark circuits in the \textit{Skywater130nm} library show delay optimization. Mapping `\text{priority}' using the \textit{Skywater130nm} library with the baseline engine (w/o DeepCell) results in a delay of 1,813.81 ps, while DeepCell reduces this to 1,199.12 ps, a 33.89\% reduction.
Second, the w/ DeepCell mapping engine unexpectedly achieves better area results than the baseline (\&nf). On average, the area is reduced by 9.3\% in the \textit{Skywater130nm} library. We attribute this improvement to our mapping approach, which modifies the critical path and creates more room for area optimization in the \textit{`\&nf'} command. 
Third, although DeepCell was not pretrained on circuits mapped in the \textit{Nangate45nm} library, it still predicts the cell type with over 90\% accuracy after fine-tuning on this library. This demonstrates DeepCell's flexibility and transferability across different technology libraries. As a result, DeepCell achieves an overall ADP reduction of 15.02\% in the \textit{Skywater130nm} library and 16.00\% in the \textit{Nangate45nm} library.


%% file: 05_conclusion.tex
\section{Conclusion}\label{Sec:Conclusion}

We introduce DeepCell, a multiview circuit representation learning framework. By integrating GNN-based encoders to learn AIG and PM representations and fusing them in a self-supervised manner, DeepCell captures richer and more generalizable embeddings. DeepCell framework introduces a novel AI-driven EDA methodology particular for the key cross-stage or cross-view applications, an area that has not been fully addressed by previous approaches. Our demonstration of DeepCell in functional ECO achieves significant reductions in patch generation costs, gate counts, and runtime. Additionally, DeepCell shows strong performance in technology mapping, providing high-quality solutions with reduced area and delay. We believe that DeepCell’s ability to fuse diverse circuit representations will unlock new possibilities for more efficient and accurate chip design.
